\documentclass[journal]{IEEEtran}

\usepackage{cite}
\usepackage{authblk}
\usepackage{graphicx}
\usepackage{svg}
\usepackage{hyperref}
\usepackage{caption}
\usepackage{subcaption}
\usepackage{tabularx}
\usepackage{outlines}
\usepackage{multirow}
\usepackage{colortbl}
\usepackage{xcolor}

\definecolor{myblue}{RGB}{0,52,120}

\begin{document}

\title{From Pixels to Planes: Minimum Ground Sample Distance for Aircraft}

\author[]{Matthew Ciolino}
\author[]{Willie Maddox}
\affil[]{PeopleTec, Inc., Huntsville, Alabama, USA}
\affil[]{\footnotesize matt.ciolino / willie.maddox \texttt{@peopletec.com}}

\maketitle

\begin{abstract}
This study investigates the impact of ground sample distance (GSD) on the detection performance of various sized aircraft using the proprietary AllPlanes 120 dataset. The data set comprises 120 civilian, military and museum aircraft from multiple satellite/aerial sources collected over two years. Resolutions ranging from 2.4 to 0.3 meters GSD were simulated. Performance metrics were derived from a YOLOv8s model trained on down-sampled versions of zoom level 19 (0.3m GSD) imagery. The results indicate that a GSD of at least 0.86m is required to accurately detect most aircraft, particularly those with wingspans shorter than 20 meters. Due to weight constraints in high-altitude platforms, this GSD specification can inform camera design to minimize weight while maintaining detection accuracy.
\end{abstract}


\IEEEpeerreviewmaketitle

\section{Introduction}
Object detection in remote sensing has been extensively reviewed, highlighting advancements in optical methods, deep learning techniques, and their impact on automated analysis tasks \cite{cheng2016survey}. This ablation study seeks to determine the minimum Ground Sample Distance (GSD) required for effective aircraft detection, motivated by the goal of minimizing optical system weight on aerial platforms. The AllPlanes 120 dataset was utilized to simulate various levels of GSD, from 2.4m to 0.3m, and to examine their impact on detection performance.

\begin{figure*}[!btp]
  \centering
  \includegraphics[width=0.98\textwidth]{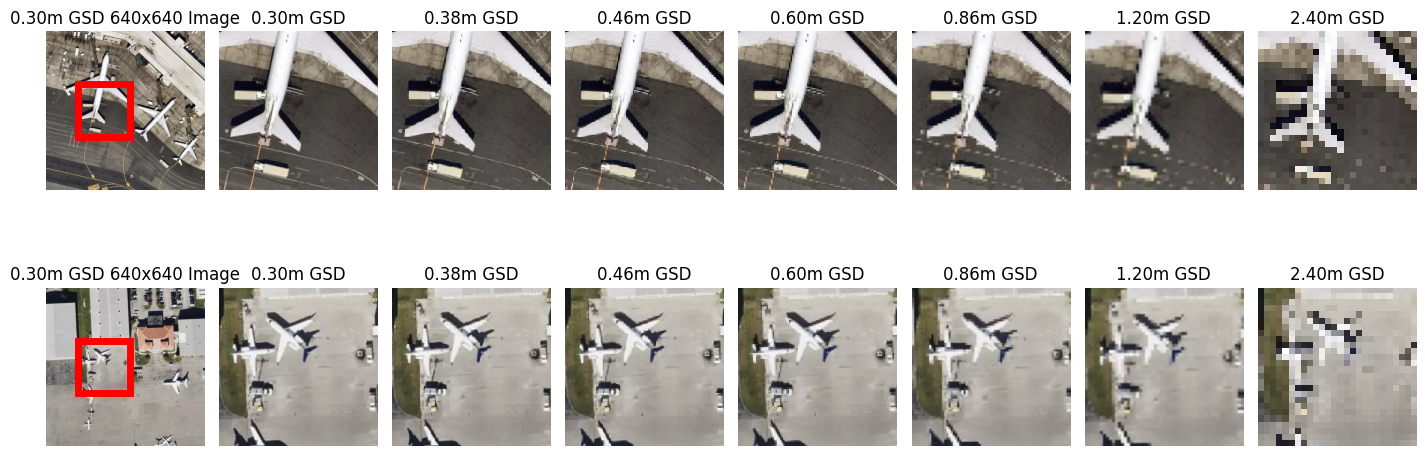}
  \caption{Resolution Comparison: 0.3m GSD Imagery at 640px $\rightarrow$ 512px, 416px, 320px, 224px, 160px, 80px}
  \label{fig:images}
\end{figure*}

\section{Dataset}
The AllPlanes 120 dataset, used for aircraft detection from satellite imagery, consists of 8,123 satellite images containing 46,459 labeled aircraft instances [Table \ref{table:dataset}]. This imagery, collected over a two-year span from 2022 to 2024, covers areas of interest such as aircraft graveyards, manufacturing facilities, military bases, airports, and museums around the world. Sources for the dataset include \textbf{Mapbox, Bing, and Google} \cite{googlemaps, bingmaps, mapsmapbox}. Aircraft are identified by ICAO Code or a grouped alias of similar codes.

\begin{table}[!ht]
    \centering
    \caption{AllPlanes 120 Dataset Statistics}
    \begin{tabular}{||l|l|l|l|l|l||}
    \hline
        \textbf{Format} & \textbf{Images} & \textbf{Instances} & \textbf{Resolution} & \textbf{Classes} & \textbf{Size} \\ \hline
        Polygon & 8,123 & 46,459 & 0.3m & 120 & 640px \\ \hline
    \end{tabular}
    \label{table:dataset}
\end{table}

\section{Methodology}

\subsection{Data Preprocessing}
The dataset consists of images of size 640x640 pixels. These were down-sampled to resolutions of 80x80, 160x160, 224x224, 320x320, 416x416, and 512x512 pixels (simulating 2.4m, 1.2m, .86m, .60m, .46m, and .38m GSD respectively) using nearest-neighbor interpolation [Figure \ref{fig:images}]. The downsampled images were used to train a YOLOv8s \cite{yolov8_ultralytics} model, configured with an imgsz of 640, batch size of 64, and trained over 50 epochs. Standard hyperparameters and data augmentation techniques were applied. The 4 keypoints were converted into a horizontal bounding box for training.

\subsection{Model Training and Evaluation}
The YOLOv8s models were well trained at each resolution. Due to the high number of classes in the AllPlanes 120 dataset, the classification error was the most common error type, which is typical for fine-grained classification datasets. Detection performance was measured using mean average precision (mAP 50-95), calculated for each resolution level.

\subsection{Grouping and Analysis}
Aircraft were grouped by wingspan to determine the minimum resolution needed for effective detection. This segmentation allowed for a focused evaluation of detection performance across aircraft size categories. Wingspan could also be used to determine the minimum number of pixels for the aircraft. For example, the Northrop T-38 Talon training jet has a wingspan of 7.7m and a length of 14m giving it an 18 pixel area at 2.4m GSD.

\section{Results}

\begin{figure}[th]
  \centering
  \includegraphics[width=0.48\textwidth]{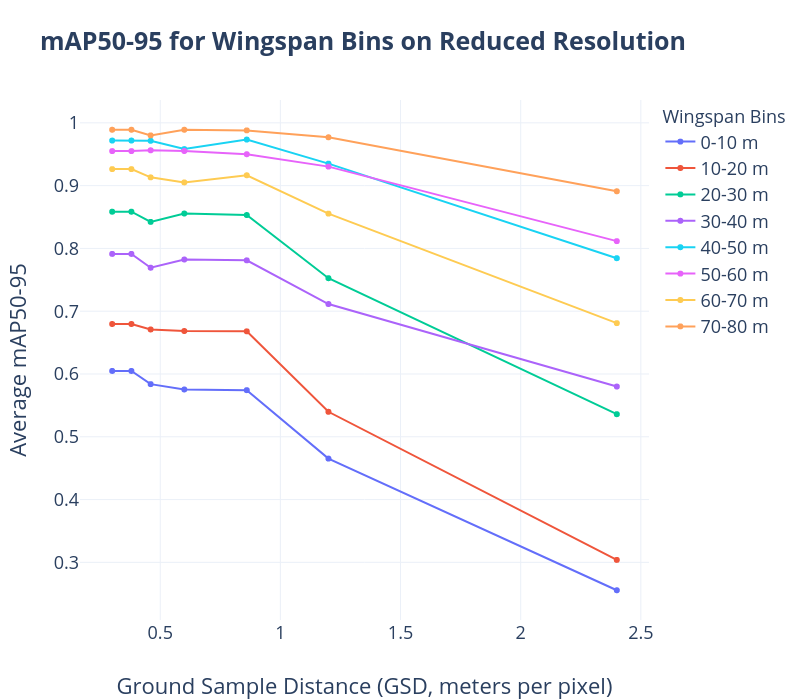}
  \caption{Binned Wingspan mAP50-95 for Reduced Resolutions}
  \label{fig:class_metrics}
\end{figure}

\subsection{Key Findings}
The analysis shows that a\textbf{ minimum GSD of 0.86m} is crucial to accurately detect most aircrafts, particularly those with wingspans below 20 meters. Performance noticeably declines at coarser resolutions, indicating that finer GSDs improve the detection of smaller targets.

\subsection{Size to Performance}
[Figure \ref{fig:class_metrics}] displays mAP50-95 for binned wingspans at various resolutions, highlighting that larger aircraft types maintain relatively high detection accuracy even at higher GSDs, while smaller aircraft classes suffer significant performance drops at resolutions coarser than 0.86m GSD. For example, classes with wingspans below 20 meters experience a substantial decrease in mAP50-95 at 1.2m GSD. Such resolution-specific variations emphasize the need for a carefully optimized GSD based on the target size distribution.

\begin{figure}[th]
  \centering
  \includegraphics[width=0.48\textwidth]{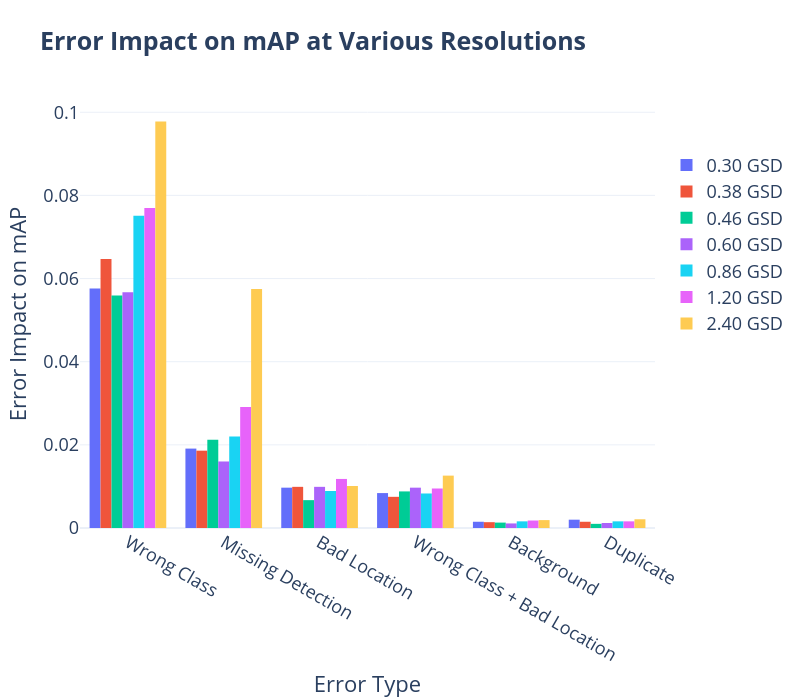}
  \caption{Error type impact on mAP for Reduced Resolutions}
  \label{fig:error}
\end{figure}

Furthermore, we investigate why the models under perform at lower resolution with an error analysis using TIDE \cite{tide-eccv2020}. We find impact of error types on overall mAP and plot their change on reduced resolutions [Figure \ref{fig:error}]. The main sources of increased error are incorrect classifications and missing detections. 

\subsection{Aircraft Examples}

Examples of an aircraft in each wingspan bin:
\begin{itemize}
    \item \textbf{~0--10m}: General Dynamics F-16 Fighting Falcon
    \item \textbf{10--20m}: Cessna 172 Skyhawk
    \item \textbf{20--30m}: Bombardier CRJ200 Regional Jet
    \item \textbf{30--40m}: Boeing 737-800 with Winglets (B738W)
    \item \textbf{40--50m}: Lockheed C-130 Hercules
    \item \textbf{50--60m}: Boeing C-17 Globemaster III
    \item \textbf{60--70m}: Boeing 777-300ER (B77W)
    \item \textbf{70--80m}: Airbus A380-800 (A388)
\end{itemize}

\section{Discussion}

\subsection{Challenges in Multi-Class Detection}
\begin{figure}[h]
  \centering
  \includegraphics[width=0.48\textwidth]{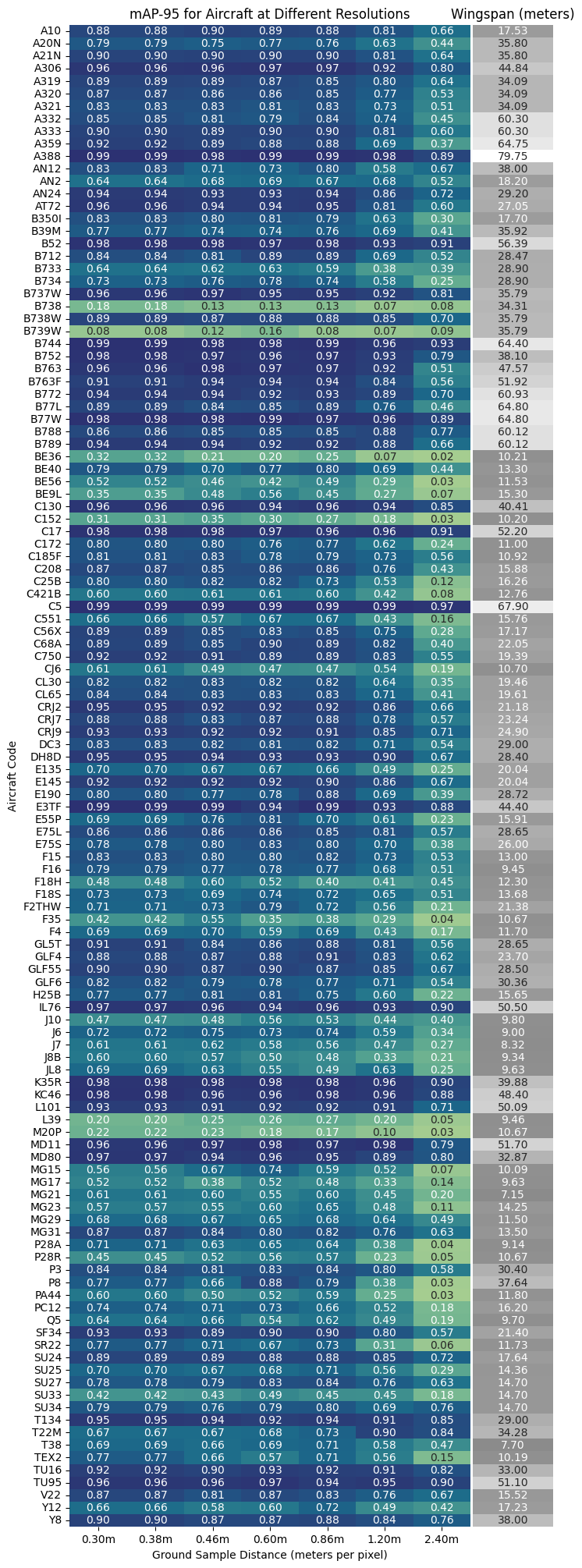}
  \caption{Per Class mAP50-95 with Loss of Resolution vs Wingspan (meters)}
  \label{fig:perf}
\end{figure}

The classification errors observed (box loss: 0.4, cls loss: 0.6, dfl loss 0.83 at 50th epoch of GSD 0.3m Training) \cite{yolov8_losses}, particularly for smaller aircraft types, highlight the challenges in fine-grained aircraft detection across varied GSD levels. The high class count within the AllPlanes 120 dataset compounded these errors, as some classes, especially those with overlapping visual features, had lower mAP50-95 scores. Addressing this could involve refining the dataset or enhancing the model's discriminative ability for visually similar classes. [Figure \ref{fig:perf}] shows each class's mAP50-95 for various resolutions.

\subsection{Technological Advances in GSD Optimization}
Advancements in sensor technologies, such as adaptive optics and high-resolution satellite imaging, have the potential to dynamically optimize GSD during image acquisition. This could allow systems to achieve higher resolution for smaller objects without imposing significant weight or energy penalties on aerial platforms. Additionally, machine learning-based image enhancement techniques, such as super-resolution networks \cite{ciolino2020training, dong2015image, wang2022remote} could mitigate resolution loss by reconstructing finer details from lower-resolution imagery, particularly for smaller aircraft.

\subsection{Limitations and Future Work}
The study focused on the YOLOv8s model, which showed significant robustness but struggled to distinguish between certain classes at lower resolutions. Future research could explore a broader range of models and consider environmental factors such as lighting and weather conditions that might impact detection performance in real-world scenarios. Higher resolution images increase file sizes, which can strain bandwidth and slow communication with ground stations, emphasizing the need for a balanced approach to system development \cite{ciolino2022enhancing}.

\section{Conclusion}
Our study concludes that a GSD of 0.86m is necessary for effective detection of smaller aircraft, emphasizing its role in reconnaissance missions where target size varies. For missions requiring trade-offs, adhering to this GSD threshold may reduce camera weight while ensuring sufficient detection accuracy, supporting lightweight, high-altitude platform missions. As always, having high-resolution imagery is generally an advantage for downstream tasks.

\bibliographystyle{./bibtex/IEEEtran}
\bibliography{./IEEEexample}

\end{document}